\documentclass[runningheads]{llncs}
\usepackage[T1]{fontenc}
\usepackage{bbding}
\usepackage{xcolor}
\usepackage{graphicx}
\usepackage{booktabs}
\usepackage{float}
\begin{document}
\title{Mask-ControlNet: Higher-Quality Image Generation with An Additional Mask Prompt}
\titlerunning{Mask-ControlNet}
%
\author{Zhiqi Huang\inst{1} \and
Huixin Xiong\inst{2} \and
Haoyu Wang\inst{1} \and Longguang Wang\inst{3} \and Zhiheng Li\inst{1}}
%
\authorrunning{Z. Huang et al.}
%
\institute{Tsinghua University, Beijing, China
\and
Megvii, Beijing, China 
\and 
Sun Yat-sen University, Guangzhou, China
}

%
\maketitle              
\begin{abstract}
Text-to-image generation has witnessed great progress, especially with the recent advancements in diffusion models. Since texts cannot provide detailed conditions like object appearance, reference images are usually leveraged for the control of objects in the generated images. However, existing methods still suffer limited accuracy when the relationship between the foreground and background is complicated. 
To address this issue, we develop a framework termed Mask-ControlNet by introducing an additional mask prompt. Specifically, we first employ large vision models to obtain masks to segment the objects of interest in the reference image. Then, the object images are employed as additional prompts to facilitate the diffusion model to better understand the relationship between foreground and background regions during image generation. Experiments show that the mask prompts enhance the controllability of the diffusion model to maintain higher fidelity to the reference image while achieving better image quality. Comparison with previous text-to-image generation methods demonstrates our method's superior quantitative and qualitative performance on the benchmark datasets.


\end{abstract}
\section{Introduction}

Recently, text-to-image generation has achieved remarkable progress with numerous models being developed, ranging from GAN~\cite{goodfellow2020generative}, VQGAN~\cite{esser2021taming} and DALL-E~\cite{reddy2021dall} to current diffusion models~\cite{sohl2015deep,peebles2023scalable}.
Despite their promising results, text-to-image models still suffer limited capability in fine-grained control of the synthetic images since text prompts cannot provide details like the object's appearance and spatial layout. To remedy this, extensive studies have been conducted to control text-to-image models using additional prompts like spatial masks \cite{avrahami2023blended}, image editing instructions \cite{brooks2023instructpix2pix,kumari2023multi,kawar2023imagic}, and other formats \cite{liu2023more,gal2022image,zhou2023shifted}.

\begin{figure}[tb]
  \centering
  \includegraphics[height=7.5cm]{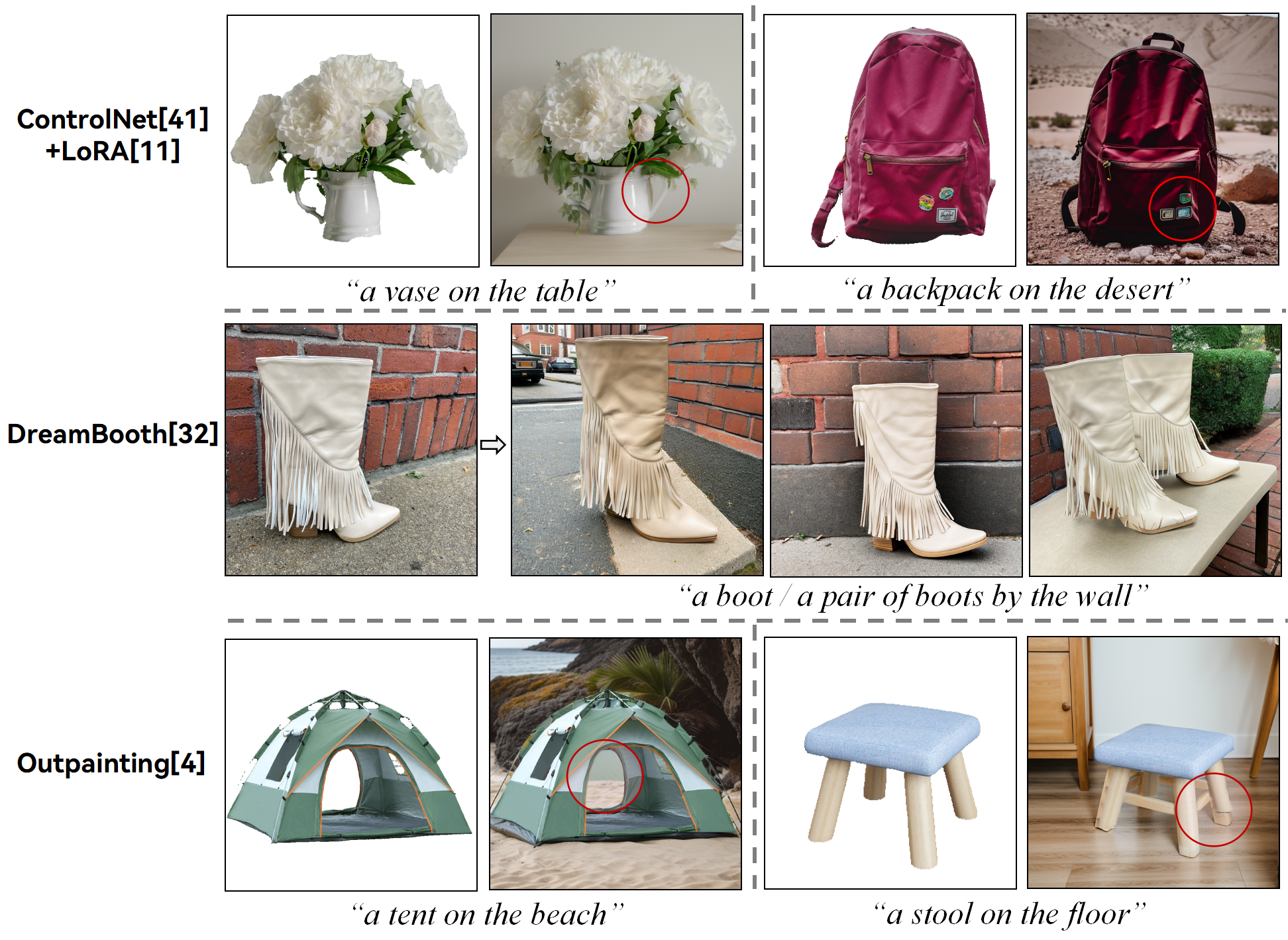}
  \caption{Limitations of existing image generation methods. The synthetic images suffer from object distortion (the first row), background overfitting (the second row) and foreground-background inharmony (the last row).
  }
  \label{fig:contrast}
\end{figure}

In numerous real-world applications, mimicking the appearance of objects in a reference image while changing the composition and contexts is under great demand. To tame text-to-image models to meet this requirement, a reference image is provided as an additional prompt\cite{ruiz2023dreambooth}. Despite the synthetic images can fit the provided prompts, they still suffer from three major limitations. 
\textbf{(i) Object distortion.} The object in the reference image may not be faithfully transferred to the synthetic image and usually suffers detail losses (the first row of Fig.~\ref{fig:contrast}).
\textbf{(ii) Background overfitting.} The concurrence of the foreground object and its background in the training set (e.g., the table in the second row of Fig.~\ref{fig:contrast}) may be overfitted. As a result, the background may also be synthesized in the generated images. 
\textbf{(iii) Foreground-background inharmony.} {Despite high object fidelity, the generated image may suffer inharmonious foreground and background (the third row of Fig.~\ref{fig:contrast}).}
Overall, these limitations are attributed to the complicated relationship between the foreground and background, which is not well modeled by the generative model.

Intuitively, the foreground and background in the prompt image play different roles during image synthesis. The foreground provides the appearance details of the object while the background helps the network to understand the context of the object and produce harmonious ones in the synthetic image. 
To better model the relationship between foreground and background, we propose to decouple these two components using an additional mask prompt. With the great advancements of recent large vision models like SAM \cite{kirillov2023segment}, the foreground object mask can be easily obtained. Specifically, the reference image is first fed to SAM to produce a mask to segment the object region. Then, the resultant image is concatenated with the reference image as the conditional information for image synthesis. The additional mask prompt facilitates the network to better maintain the object details and model the foreground-background relationship, resulting in higher-quality synthetic images. 

The main contributions are summarized as below:

\begin{itemize}
\item[1)] We propose a framework termed Mask-ControlNet to achieve higher-quality image generation by introducing an additional mask prompt. With the help of this mask prompt, the foreground and background in the reference image can be decoupled and well-modeled to improve the quality of the synthetic image.


\item[2)] We conduct extensive experiments to study our framework. Quantitative and qualitative results show that our framework is able to generate high-quality images with fewer artifacts.

\end{itemize}

\section{Related Work}
In this section, we briefly review recent advances in text-to-image generative models and controllable generative models.

\subsubsection{Text-to-Image Generative Models } 
Text-to-image generative models have been studied for decades, with diverse networks being developed, including
VAE ~\cite{kingma2013auto}, PixelCNN~\cite{van2016conditional,van2016pixel}, Glow~\cite{kingma2018glow}, and GAN~\cite{goodfellow2020generative}. Among these methods, GAN stands out as one of the most popular models. Over the past five years, GANs have witnessed substantial progress and are capable of generating high-resolution images up to $1024\times1024$ pixels or higher~\cite{karras2017progressive,brock2018large,kang2023scaling}. 
Despite the theoretical equilibrium, GAN’s training is still faced with challenges like instability and mode collapse.

The advent of diffusion models~\cite{sohl2015deep,peebles2023scalable} offers a stable, high-quality, and controllable solution. DDPM~\cite{ho2020denoising} surpassed GAN using a U-net denoising autoencoder. DDIM~\cite{song2020denoising} further improves image synthesis with implicit probabilistic models and continuous-time dynamics. Recently, stable diffusion model further outperforms DDIM by introducing a latent diffusion model~\cite{rombach2022high} to replace pixel-level diffusion models and using an autoregressive flow model to realize the prior distribution of the latent space. 

Diffusion models usually introduce interfaces when conducting fine-grained image editing such as color and texture details~\cite{kumari2023multi,kumari2023ablating,kawar2023imagic}. To remedy this, Texture Inversion~\cite{gal2022image} personalizes the generated content using a small set of images under the same overarching theme.  DreamBooth~\cite{ruiz2023dreambooth} associates target themes with unique identifiers, enabling these identifiers to be embedded in the model's output domain, which synthesizes realistic images of specific themes in differentiated scenarios.

\subsubsection{Controllable Generative Models}
Despite high quality, diffusion models commonly suffer inferior fidelity to the conditional image. To improve the controllability of diffusion models, more complicated texts are employed to provide detailed descriptions of the image. Liu et al.~\cite{liu2023more} proposed a semantic diffusion-guided framework that utilizes CLIP~\cite{radford2021learning} or other image-matching models to compute the similarity between the text prompt and the generated images to guide the diffusion model in generating images that better match the semantic requirements. Avrahami et al.~\cite{avrahami2023blended} introduced a text-driven image editing method that allows the modification of specific shards in an image according to user-provided text. 

As the text cannot provide fine-grained control (e.g., object appearance) to the generated images, the results may not meet the requirements of users. To address this issue, ControlNet~\cite{zhang2023adding} extends text-to-image diffusion models with conditional control~\cite{mou2023t2i,li2023gligen,huang2023composer,ju2023humansd,voynov2023sketch,zheng2023layoutdiffusion}, allowing the generation of images that match diverse types of user-specified prompts. Outpainting is proposed to preserve the essential components in a reference image while extending or enhancing the background. 
Subsequent methods~\cite{lugmayr2022repaint,corneanu2024latentpaint,xie2023smartbrush} use a pre-trained unconditional diffusion model as a generative prior, and then achieve conditional generation by sampling randomly shaped unmasked regions during the reverse diffusion process. While these methods perform well on simple objects, they cannot well handle complicated scenes and encounter challenges including object distortion, background overfitting, and foreground-background inharmony.


\section{Methodology}

Given a reference image of a specific object, our objective is to generate images that maintain high detail fidelity of the object while synthesizing diverse contexts and compositions conditioned on the text prompts.

\subsection{Training-time Framework}
\begin{figure}[t]
  \centering
  \includegraphics[height=6cm]{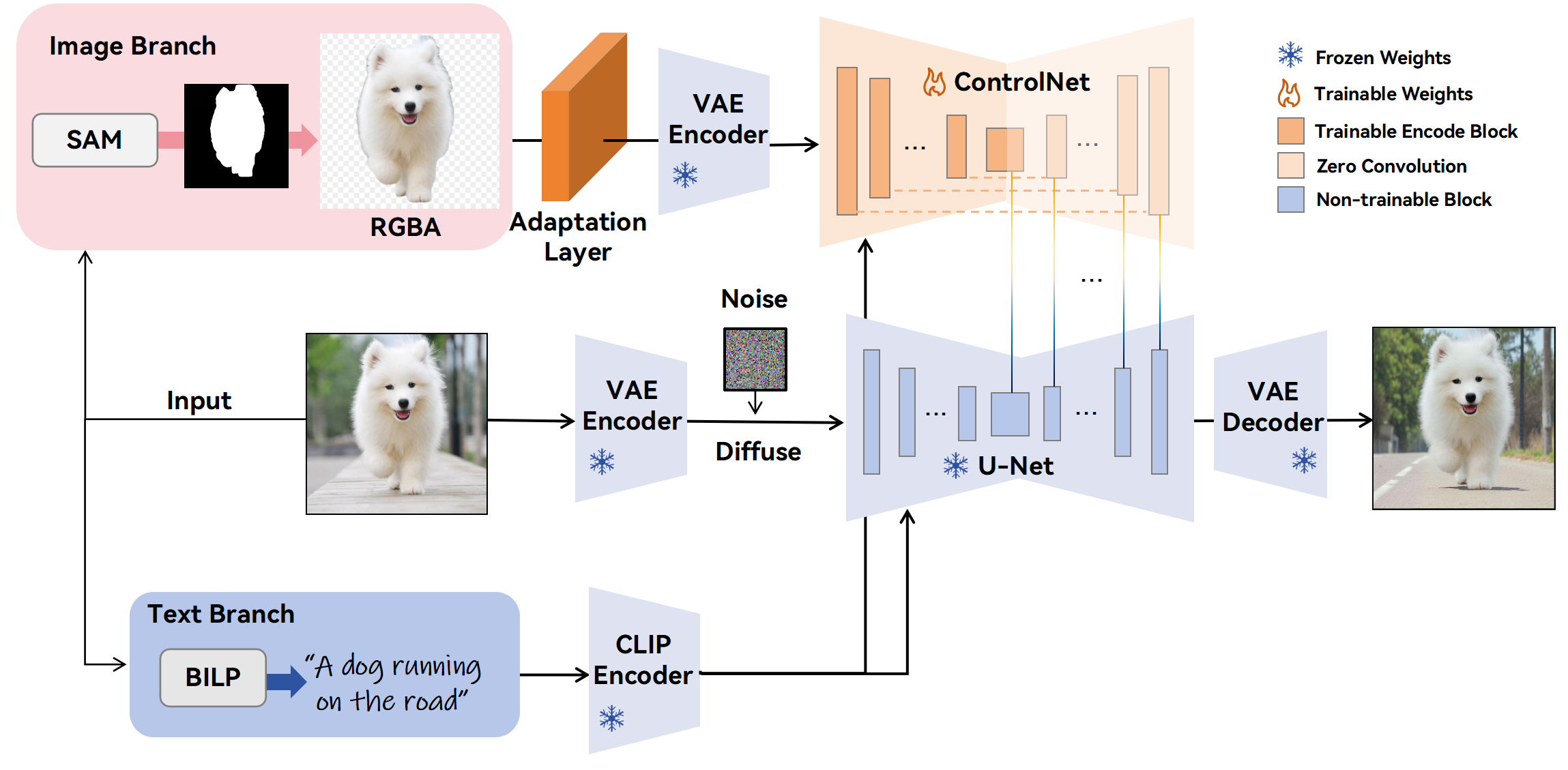}
  \caption{An illustration of our framework during the training phase. 
  }
  \label{fig:training}
\end{figure}

As shown in Fig.~\ref{fig:training}, our framework is built on top of a diffusion model and is trained in a self-supervised manner. First, the input image is fed to the VAE encoder to obtain feature maps $F$ and then the noise is progressively added, resulting in $F_t$. Here, $t$ represents the number of times noise is added. Afterward, the noisy feature maps $F_t$ are passed to the diffusion model to predict the noise and reconstruct the input image. 

In parallel to the main path, our framework has an image branch and a text branch to provide additional conditions for the diffusion model. In the image branch, the input image is first fed to SAM~\cite{kirillov2023segment} to produce the object mask. Then, the resultant mask is used to segment the object in the image. Next, the concatenation of the object image and the image is passed to an adapter layer. Afterward, a VAE encoder and ControlNet are employed to control the diffusion model to reconstruct the input image. In the text branch, BLIP\cite{pmlr-v162-li22n} is adopted to extract textual descriptions of the input image. Then, the extracted text prompt is fed to CLIP to provide additional control to the diffusion model. 
Following \cite{zhang2023adding}, the features extracted from the text and image prompts are connected to the diffusion model with zero convolution layers.

During optimization, only the adapter layer and the ControlNet are trainable while the diffusion model is frozen. The loss function used is defined as:

\begin{equation}
L=E_{z_{0},t,c_{t},c_{f},\epsilon\sim N(0,1) }||\epsilon-\epsilon_{\theta }(z_{t},t,c_{t},c_{f})||_{2}^2, 
\label{eq:loss}
\end{equation}
{where \(z_{0}\) represents the data in the latent space, \(c_{t}\) and \(c_{f}\) are the text condition and the latent condition, respectively.}

\subsection{Inference-time Framework}
\begin{figure}[t]
  \centering
  \includegraphics[height=5cm]{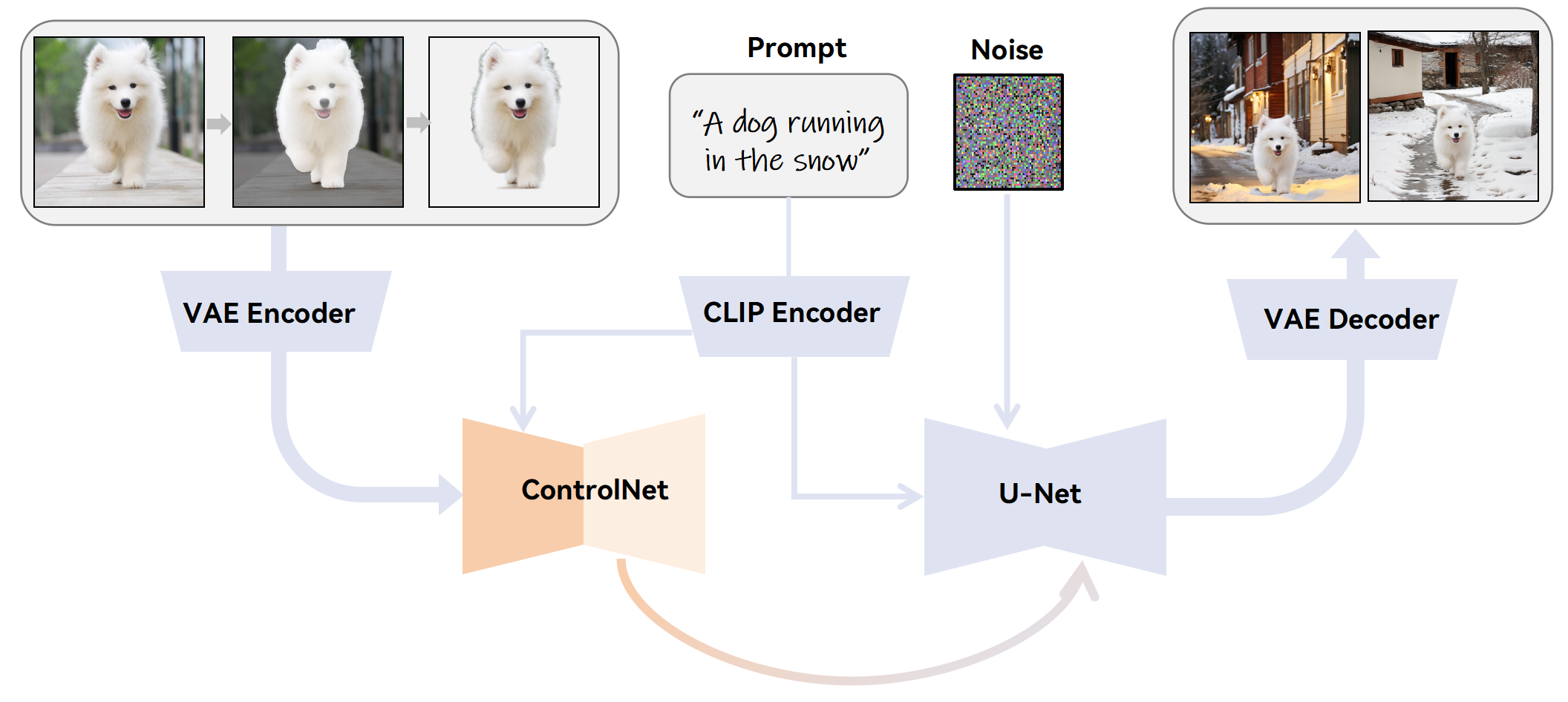}
  \caption{An illustration of our framework during the inference phase. }
  \label{fig:inference}
\end{figure}

Our framework during inference time is illustrated in \ref{fig:inference}. First, the reference image is fed to SAM to produce a mask to segment the object. Then, the concatenation of the object image and the reference image is passed to the VAE encoder. Meanwhile, the text that describes the context of the generated image is fed to CLIP. Next, the features extracted from the image and text prompts are passed through ControlNet and used as conditions for the diffusion model to synthesize an image from a noise image.

\section{Experiments}
In this section, we first introduce the experimental setup. Then, we conduct experiments to compare our method with previous approaches. Next, we conduct several ablation studies to demonstrate the effectiveness and robustness of the mask.

\subsection{Experimental Setup}
During the training phase, we collected 100,000 images from numerous websites using keywords such as people, food, animals, furniture, cosmetics, automobiles, and clothing. {In addition, we selected 100,000 images from the SA1B and COCO datasets.}  After data cleaning and annotation, a total of 130,000 valid images and approximately 300,000 valid masks are obtained as the training set.

During the test phase, the DreamBooth dataset is included for evaluation, which comprises 30 categories like backpacks, toys, dogs, cats, and sunglasses. Each category has about 5-8 images.
FID \cite{seitzer2020pytorch}, PSNR~\cite{tanchenko2014visual}, SSIM~\cite{wang2004image} and LPIPS~\cite{zhang2018unreasonable} are employed to evaluate  the generated images. In addition, to measure the object grounding accuracy from the reference image and the text to the generated image, we adopt the CLIP score~\cite{radford2021learning} and DINO score~\cite{caron2021emerging}.

\subsection{Performance Evaluation}

\vspace{0.1cm}
\noindent\textbf{(1) Quantitative Results}
\vspace{0.1cm}

We compare our method with three representative methods, including DreamBooth~\cite{ruiz2023dreambooth}, ControlNet~\cite{zhang2023adding}+LoRA~\cite{hu2021lora}, and Outpainting~\cite{Cai_2023_ICCV}. These methods are capable of generating images with different backgrounds for specified objects. {DINO and CLIP-I are used to evaluate the object fidelity while CLIP-T is employed to calculate prompt fidelity. As shown in \ref{tab:matric2}, our method is more faithful to the text prompts with higher CLIP-T scores. Additionally, our method outperforms other methods except ControlNet+LoRA in terms of both CLIP-I and DINO. It should be noted that ControlNet+LoRA requires partial input images for fine-tuning. Although our method does not conduct this fine-tuning process, competitive results are produced, which demonstrates the effectiveness of our method. As compared to DreamBooth, our method achieves much better performance in terms of all metrics.}

\begin{table}[t]
  \caption{Prompt fidelity comparison of images produced by different methods.}
  \label{tab:matric2}
  \centering
  \setlength{\tabcolsep}{5pt} 
  \begin{tabular}{@{}llll@{}}
    \toprule
    Method & CLIP-T$\uparrow$ & CLIP-I$\uparrow$ & DINO$\uparrow$\\
    \midrule
    DreamBooth~\cite{ruiz2023dreambooth} & 0.171 & 0.828 & 0.527 \\
    ControlNet~\cite{zhang2023adding}+LoRA~\cite{hu2021lora} & \underline{0.173} & {\bf0.873} & {\bf0.607}\\
    Outpainting~\cite{Cai_2023_ICCV} & 0.166 & 0.849 & 0.423\\
    Mask-ControlNet & {\bf0.175} & \underline{0.858} & \underline{0.593}\\
    Real Images & 0.176 & 0.885 & 0.774 \\
  \bottomrule
  \end{tabular}
\end{table}

\begin{table}[t]
  \caption{Quantitative results achieved by ControlNet+LoRA and Mask-ControlNet. }
  \label{tab:matric1}
  \centering
  \setlength{\tabcolsep}{5pt} 
  \begin{tabular}{@{}lllll@{}}
    \toprule
    Method & FID$\downarrow$ & PSNR$\uparrow$ & SSIM$\uparrow$ & LPIPS$\downarrow$\\
    \midrule
    ControlNet~\cite{zhang2023adding}+LoRA~\cite{hu2021lora} & 6.161 & 28.06 & 0.957 & 0.031\\
    Mask-ControlNet &  {\bf5.172} &  {\bf30.67} & \textbf{0.958} &\textbf{0.022}\\
    Real Images & 0.000 & INF & 1.0 & 0.00 \\
  \bottomrule
  \end{tabular}
\end{table}
{Then, we compare the quality of images synthesized by different methods in \ref{tab:matric1}. As we can see, our Mask-ControlNet surpasses ControlNet+LoRA by notable margins on all metrics. Specifically, our Mask-ControlNet demonstrates superior performance in terms of FID and PSNR (5.172 and 30.67, respectively) as compared to ControlNet+LoRA (6.161 and 28.06). This indicates that our method generates images that closely match the distribution of real images and better match the ground truths in terms of pixel-level fidelity.}

\vspace{0.1cm}
\noindent\textbf{(2) User Study}
\vspace{0.1cm}

We further invited users to evaluate the aesthetics, grounding accuracy, and the realness of the images generated by different methods.  Average Human Ranking (AHR) is employed to rank each result on a scale of 1 to 5 (the higher the better). We asked 207 users to answer questionnaires of 20 comparative questions. They were rated for a set of images of the same item generated by different methods. Totaling 4,140 answers were collected and 33.8\% of users had no understanding of image generation models. The results are presented in \ref{tab:matric3}. 
\begin{table}[H]
  \caption{Average User Ranking (AUR) of image quality in terms of aesthetics, accuracy and realness. 1 to 5 indicates worst to best.}
  \label{tab:matric3}
  \centering
  \setlength{\tabcolsep}{5pt} 
  \begin{tabular}{@{}llll@{}}
    \toprule
    Method & Aesthetics$\uparrow$ & Accuracy$\uparrow$ & Realness(\%)$\uparrow$\\
    \midrule
    DreamBooth~\cite{ruiz2023dreambooth} & 3.909 & 3.299 &15.59\\
    ControlNet~\cite{zhang2023adding}+LoRA~\cite{hu2021lora} & 3.545 & 3.888 & 32.46\\
    Outpainting~\cite{Cai_2023_ICCV} & 3.883 & 3.480 & 18.19\\
    Mask-ControlNet & \textbf{4.377}& \textbf{4.330 }& \textbf{61.04} \\
  \bottomrule
  \end{tabular}
\end{table}
As we can see, our method outperforms other methods by notable margins in terms of all metrics. Particularly, the results synthesized by our method achieve significantly higher accuracy (4.330 vs. 3.888). This demonstrates that our method can better maintain the object details in the reference images with higher fidelity. 
In addition, 61.04\% of the users cannot distinguish between our results and real images. This further validates the effectiveness of our method to produce images high in realness.

\vspace{0.1cm}
\noindent\textbf{(3) Qualitative Analysis}
\vspace{0.1cm}

We further compare the visual quality of the images produced by different methods. We focus on the three challenges as presented in Sec.1.

\vspace{0.1cm}
\noindent\textbf{(i) Object Distortions}
\vspace{0.1cm}

\begin{figure}[tb]
  \centering
  \includegraphics[width=0.85\linewidth]{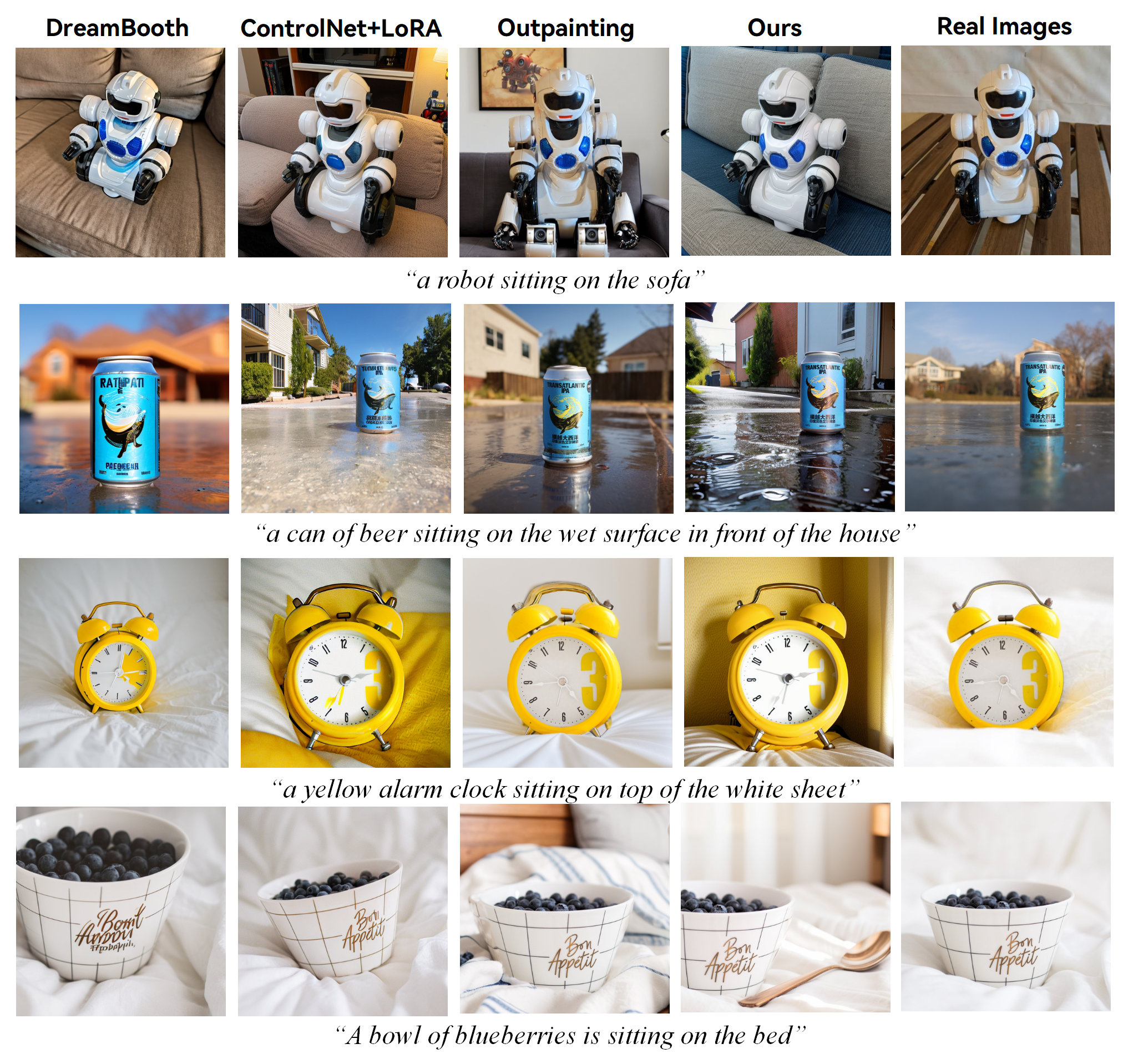}
  \caption{Synthetic images produced by different methods under the same prompt. From the perspectives of object edges, texture, text, color, etc., our method generates images that are more lossless and closer to reality.
    }
  \label{fig:imgcontr}
\end{figure}

In Fig.~\ref{fig:imgcontr}, it is evident that the results produced by other methods suffer object distortions. For example, {in the second row, DreamBooth and ControlNet+LoRA generate incorrect text on the can, while Outpainting produces a distorted bottom. } In contrast, our method maintains higher fidelity of the objects and better preserves their details in the generated images.
\begin{figure}[tb]
  \centering
   \includegraphics[width=0.8\linewidth]{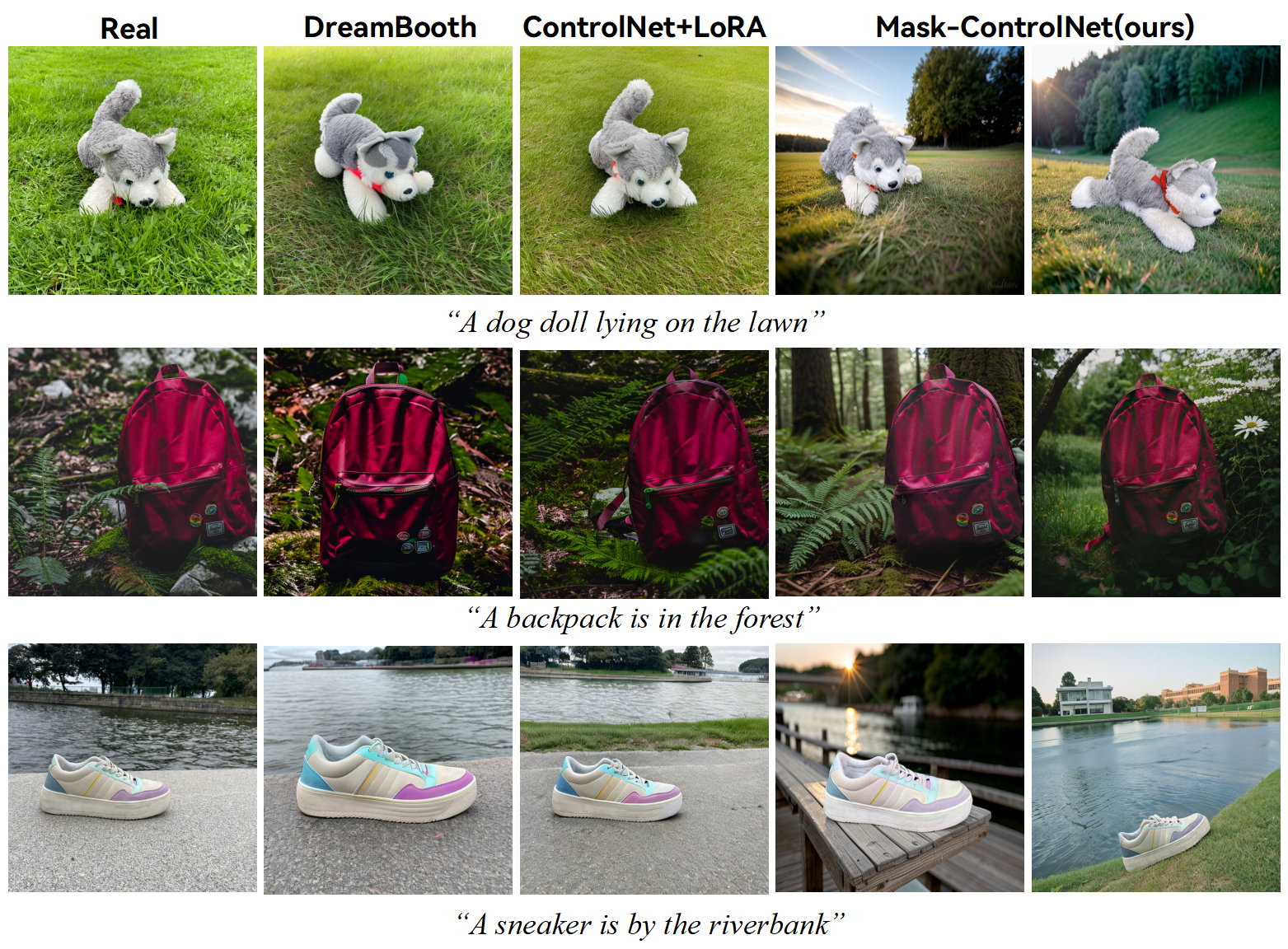}
   \caption{Background contrast generated under the same prompt. From the figure, it can be seen that our method can generate more diverse backgrounds.}
   \label{fig:wolf}
\end{figure}
\begin{figure}[tb]
  \centering
   \includegraphics[width=1.0\linewidth]{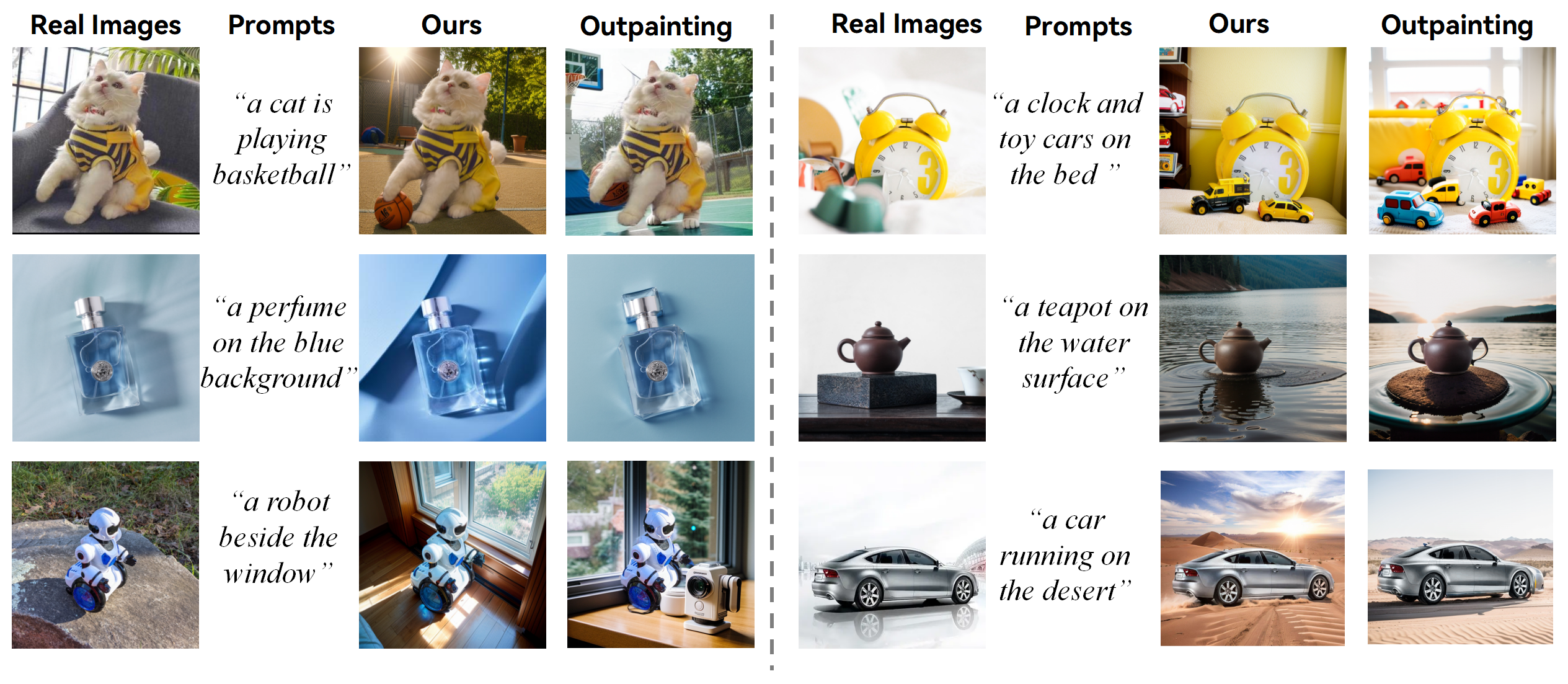}
   \caption{Comparison of foreground-background harmony in different scenarios. The images from upper left to lower right present six different interaction scenarios, including contact, refraction, shadow, obstruction, reflection, and dynamic effects.}
   \label{fig:reflect}
\end{figure}

\vspace{0.1cm}
\noindent\textbf{(ii) Background Overfitting}
\vspace{0.1cm}

In Fig.~\ref{fig:wolf}, we can see that other methods are prone to introducing the background of the object in the reference image to the synthetic images, such as the riverbank in the third row. In contrast, our method can better understand the relationship between foreground and background, thereby generating images with diverse contexts.

\vspace{0.1cm}
\noindent\textbf{(iii) Foreground-background Inharmony}
\vspace{0.1cm}

In Fig.~\ref{fig:reflect}, we can observe that the foreground and background are more harmonious in our results as compared to other methods. For example, in the first scene (a cat is playing basketball), Outpainting cannot well understand the text and generates a ball behind the cat. In contrast, our method can better understand the text and synthesizes a ball under the foot. For other challenging conditions like occlusions, reflections, and shadows, the relationship between the foreground and the background can also be well handled. These results clearly validate the superiority of the proposed method. 

\subsection{Model Analyses}
\begin{figure}[H] 
   \centering 
   \includegraphics[width=1.0\linewidth]{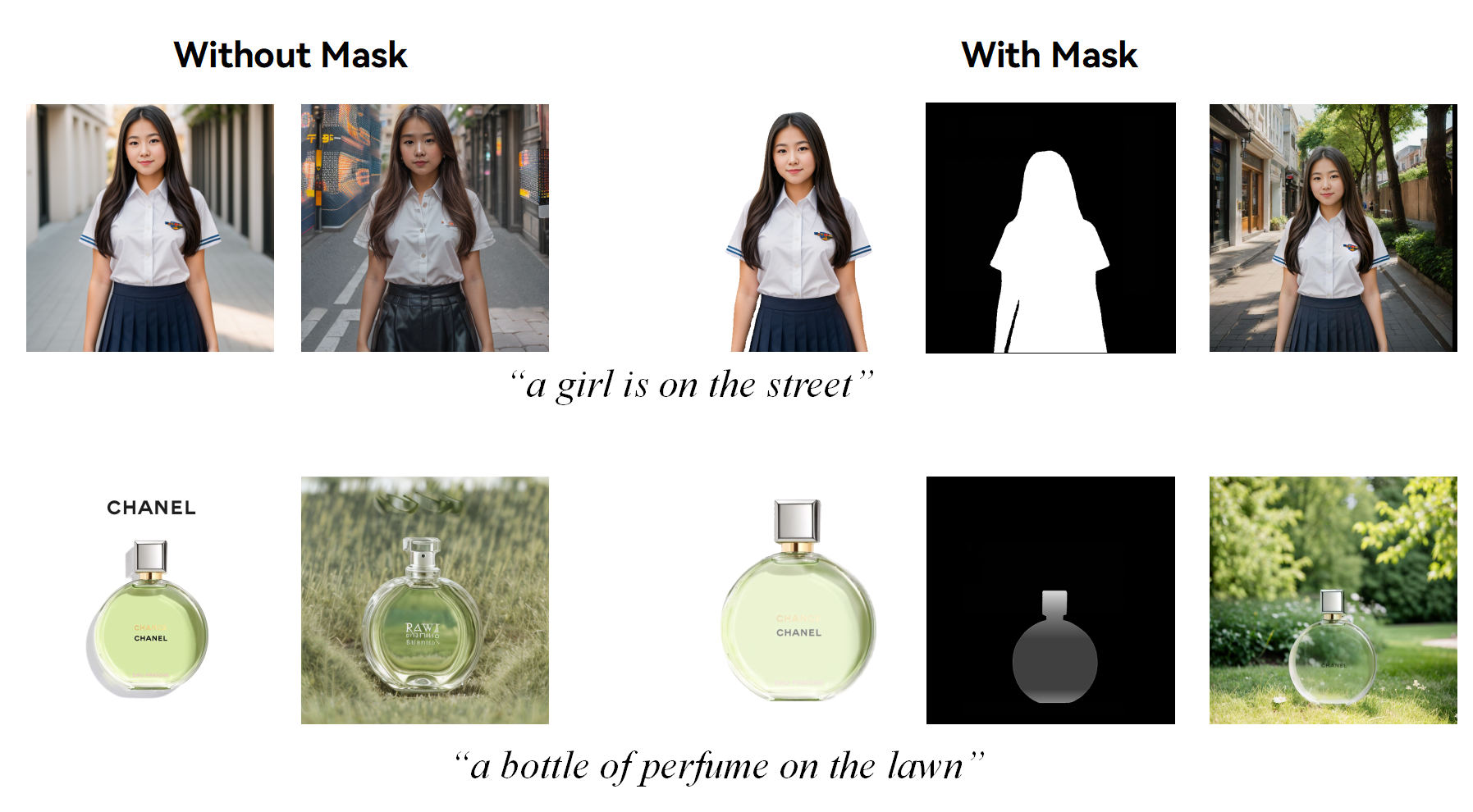} 
   \caption{Images synthesized by our method with and without masks.} 
   \label{fig:nomask} 
\end{figure}

\vspace{0.1cm}
\noindent\textbf{(1) Effectiveness of Mask Prompts}
\vspace{0.1cm}
\begin{table}[t]
  \caption{Ablation results on mask prompts.}
  \label{tab:ablation}
  \centering
  \setlength{\tabcolsep}{5pt} 
  \begin{tabular}{@{}lll@{}}
    \toprule
     Method  & CLIP-I$\uparrow$ & DINO$\uparrow$ \\
    \midrule
    w/ mask & {\bf0.858} & {\bf0.593} \\
    w/o mask & 0.812 & 0.386  \\
  \bottomrule
  \end{tabular}
\end{table}
To demonstrate the effectiveness of masks, ablation experiments are conducted. Specifically, we develop a variant that directly uses the reference image as the prompt without leveraging the mask to segment the objects. As shown in Table \ref{tab:ablation}, the mask prompt facilitates our method to better maintain the object details in the reference image to achieve higher fidelity scores in terms of both CLIP-I and DINO. {Qualitatively, as shown in \ref{fig:nomask}, the model without mask exhibits relatively low quality. In contrast, by incorporating a mask as an additional prompt, our method is able to generate images that are more faithful to the reference image (e.g., the T-shirt of the girl in the first row) with higher perceptual quality.
}

\begin{figure}[tb] 
    \centering 
    \includegraphics[width=0.85\linewidth]{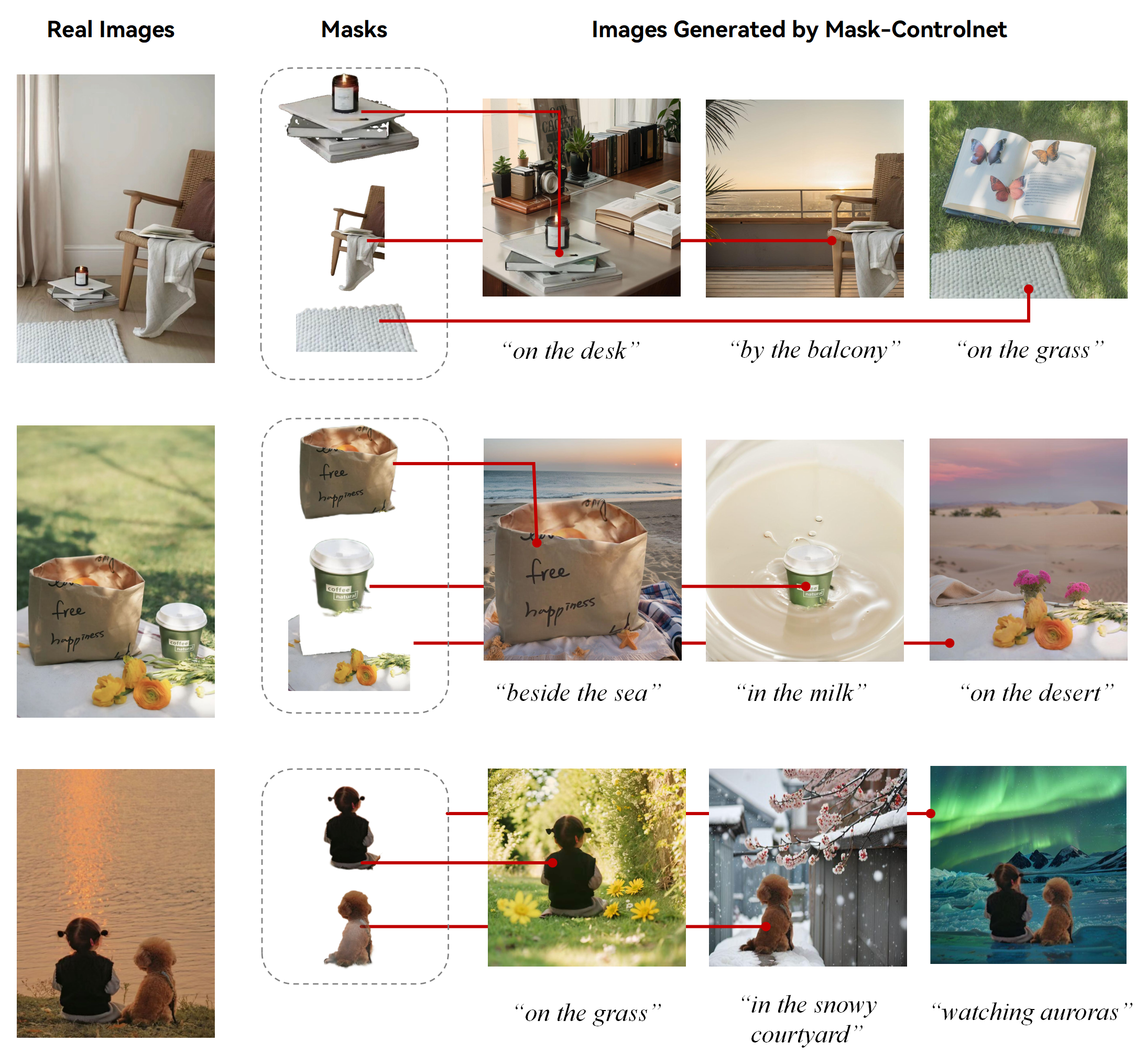} 
    \caption{Images generated using the same reference but different masks. } 
    \label{fig:role} 
\end{figure} 

\vspace{0.1cm}
\noindent\textbf{(2) Flexibility of Mask Prompts}
\vspace{0.1cm}

We further test the flexibility of the mask prompts in our framework. When an image containing multiple objects is employed as the reference image, our framework can flexibly transfer objects of interest to the generated images using different masks, as shown in Fig. \ref{fig:role}. It can be observed that our method produces visually promising images using the same reference image with different masks as conditions. Without our mask prompt, the same reference image is used as a condition. As a result, the diffusion model may be confused and cannot well understand the text prompts.

\section{Conclusion}
In this paper, we present a simple yet effective framework to synthesize high-quality images with an additional mask prompt. To better model the relationship between foreground and background, we use SAM to obtain the object mask to provide additional cues. With this addition of conditional information, the network can well capture the foreground-background correlation and generate visually more pleasing results. Extensive experiments demonstrate the effectiveness of our method and the superiority of our generated images.
\bibliographystyle{splncs04}
\bibliography{mybibliography}

%




\end{document}